\documentclass[letterpaper,10pt,conference]{ieeeconf}

\IEEEoverridecommandlockouts
\ifdefined\pdfminorversion
\fi

\ifdefined\pdfobjcompresslevel
\fi

\usepackage{graphicx}
\usepackage{amsmath,amssymb}

\usepackage{cite}

\usepackage[table]{xcolor}

\usepackage{booktabs}
\usepackage{array}
\usepackage{tabularx}
\usepackage{multirow}

\usepackage{algorithm}
\usepackage{algpseudocode}

\usepackage{url}

\graphicspath{{figures/}}
\definecolor{MethodBlue}{RGB}{45,92,136}
\definecolor{MethodOrange}{RGB}{166,92,28}
\definecolor{MethodGreen}{RGB}{55,112,78}

\definecolor{SoftGray}{RGB}{238,238,238}

\definecolor{DraftRed}{RGB}{160,35,35}

\definecolor{CitationGreen}{RGB}{0,128,0}
\definecolor{ReferenceRed}{RGB}{180,0,0}

\newcommand{\refnum}[1]{%
  \textcolor{ReferenceRed}{\ref{#1}}%
}

\newcommand{\figref}[1]{%
  Fig.~\refnum{#1}%
}

\newcommand{\tabref}[1]{%
  Table~\refnum{#1}%
}

  \newcommand{\todo}[1]{}
  \newcommand{\note}[1]{}

\title{\LARGE \bf
DynaForge: Planning-Guided Residual Learning for Dynamic Manipulation Demonstration Generation
}

  \author{Yiyang Jin$^{1*}$, Yu Zheng$^{1*}$, Xiao He$^{2}$,  and Hesheng Wang$^{1}$
    \thanks{*The first two authors contributed equally. 
    Corresponding Author: Hesheng Wang (e-mail: wanghesheng@sjtu.edu.cn).
    }%
    \thanks{$^{1}$School of Automation and Intelligent Sensing, Shanghai Jiao Tong University, Shanghai 200240, China.
    }
    \thanks{$^{2}$LandSpace Technology Co.,Ltd., China} 
    }

\ifdefined
\fi

\begin{document}
\maketitle
\thispagestyle{empty}
\pagestyle{empty}

\begin{abstract}
Dynamic object manipulation is essential for robots operating in real-world environments, yet methods for generating high-quality demonstrations remain limited. Methods designed for static tasks do not readily transfer to dynamic settings. Among dynamic demonstration generators, planning-based methods can fail near contact, while DOMINO-style replay simplifies dynamic interactions and may limit the experience available for policy learning. We present DynaForge, a planning-guided framework that learns residual corrections for dynamic manipulation demonstration generation. DynaForge combines low-frequency global planning with high-frequency object-centric inverse kinematics across task phases, and applies a residual policy to correct actions during dynamic interaction. An implicit curriculum groups rollouts under matched conditions and selects mixed-success groups, focusing residual reinforcement learning on the evolving competence frontier. On Can and Bottle, it uses 0.73x as many optimizer steps as vanilla GRPO at the same nominal environment-step budget, with higher observed final success rates. Across nine simulation tasks, DynaForge increases mean demonstration-generation success from 41.30\% of the planning prior to 78.37\%. With 800 demonstrations per task, DP3 policies trained on DynaForge data achieve 49.11\% mean success, compared with 7.07\% for DOMINO data. On three real-world dynamic tasks, DynaForge-trained policies achieve 30–60\% success, compared with 0–10\% for DOMINO-trained policies, showing the ability of DynaForge for sim-to-real transfer.
\end{abstract}

\section{Introduction}
\label{sec:introduction}

Real-world robotic manipulation rarely occurs in fully static environments~\cite{billard2019trends}. Objects may roll, slide, or otherwise move according to task-dependent temporal patterns. Robots must therefore coordinate perception, motion, and contact under changing interaction conditions~\cite{akinola2021dynamic, xie2026dynamicvla, yun2026physmani,zhang2025catch}. Despite progress in large-scale robot learning, existing datasets and automated demonstration-generation pipelines still focus primarily on static or quasi-static manipulation~\cite{o2024open,khazatsky2024droid,mandlekar2023mimicgen,chen2025robotwin}. Demonstrations capturing time-sensitive interactions between robots and moving or perturbed objects remain scarce. This scarcity limits the ability of data-driven methods, including imitation learning (IL) and vision-language-action models (VLAs), to perform dynamic manipulation~\cite{xie2026dynamicvla,yun2026physmani,liao2026dynamicmanip,lin2025data}.

\begin{figure}[tbp]
    \centering
    \includegraphics[width=\columnwidth]{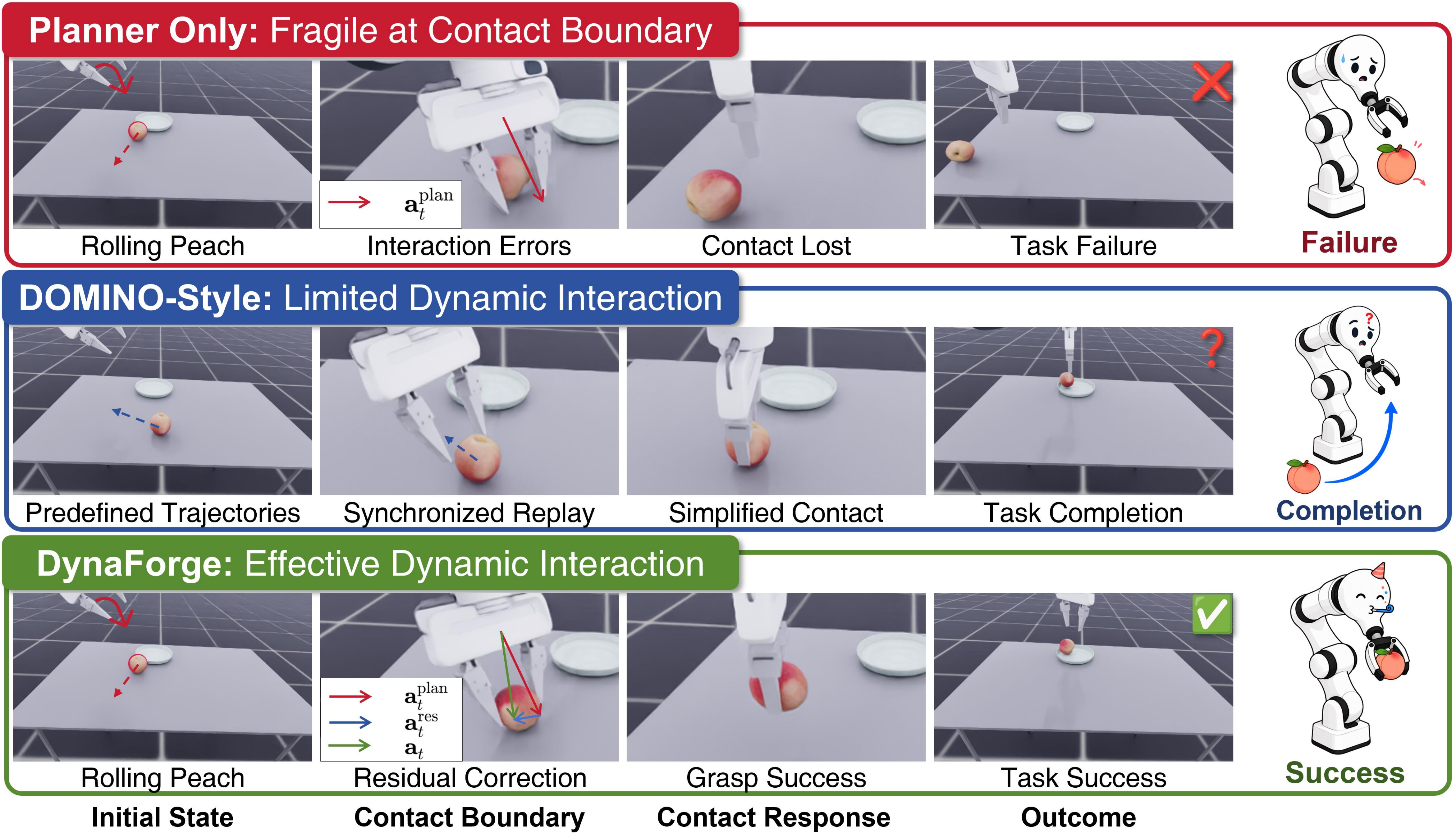}
    \caption{\textbf{Dynamic interaction at contact boundaries.} In the rolling-peach example, planner-only execution loses contact, while DOMINO-style replay completes the task with simplified dynamic interactions. DynaForge applies residual correction at the contact boundary to secure the grasp and complete the task.}
    \label{fig:motivation}
\end{figure}

Existing data collection paradigms struggle to generate dynamic manipulation demonstrations at scale. Short contact windows make manual teleoperation costly and difficult to execute reliably~\cite{tony2023aloha,xie2026dynamicvla,yun2026physmani}. Video retargeting is also limited by embodiment gaps and contact discrepancies between observed behaviors and executable robot actions~\cite{bahl2022human,shaw2023videodex,lepert2025masquerade}. Recent automated methods adapt to dynamic environments through state-machine execution~\cite{xie2026dynamicvla}, trajectory adaptation~\cite{pomponi2026dynamimicgen}, online predictive planning~\cite{yun2026physmani}, and data augmentation~\cite{liao2026dynamicmanip,pomponi2026dynamimicgen}. However, in planning- and retargeting-based pipelines with fixed execution rules, adaptation is primarily expressed as updates to targets or trajectories. These pipelines provide geometric guidance but do not learn execution-level corrections from rollout outcomes. This limitation is particularly relevant at contact boundaries, where execution errors can disrupt dynamic interaction (\figref{fig:motivation}).

We therefore treat motion planning as a structured prior for dynamic demonstration generation rather than as the complete execution mechanism. The planning system represents task phases and constructs nominal motions under collision and kinematic constraints. A residual policy learns corrections to the nominal planner action from rollout outcomes. This decomposition assigns global motion construction to planning and execution-level correction to learning.

Based on this decomposition, we propose \textbf{DynaForge}, a planning-guided residual learning method that generates dynamic manipulation demonstrations at scale through two complementary designs: \textbf{1)} DynaForge combines low-frequency motion planning with high-frequency local interaction updates. During geometrically stable transitions, the path planner is invoked infrequently to generate global collision-free motions~\cite{sundaralingam2023curobo}. Near dynamic contact boundaries, high-frequency, object-centric inverse-kinematics (IK) updates track changes in local interaction geometry. \textbf{2)} DynaForge introduces an implicit curriculum for residual learning under dynamic perturbations. We group rollouts under matched perturbation conditions and optimize group-relative objectives. Groups in which all rollouts succeed or all rollouts fail provide little comparative information and are excluded from the corresponding update. Optimization therefore focuses on groups near the current success-failure boundary. This filtering mechanism allows the effective training distribution to shift toward more challenging perturbation conditions as the residual policy improves, without requiring an explicit difficulty schedule.

Our main contributions are:

\begin{itemize}

\item We propose DynaForge, a planning-guided residual learning method for dynamic manipulation demonstration generation. DynaForge learns residual corrections near contact boundaries over a structured planning prior.

\item We introduce an implicit curriculum for residual learning through matched-perturbation group-relative learning. Excluding rollout groups with uniformly successful or failed outcomes focuses optimization near the evolving success-failure boundary without an explicit difficulty schedule.

\item We evaluate demonstration generation on nine dynamic manipulation tasks in simulation and assess the utility of the generated data by training DP3~\cite{Ze2024DP3} policies for evaluation in simulation and on three real-world tasks.

\end{itemize}

\section{Related Work}
\label{sec:related}

\subsection{Scalable Demonstration Generation}

Large-scale, high-quality demonstrations are essential for data-driven robotic manipulation. Existing generation methods mainly target static or quasi-static settings~\cite{ye2026datapyramid}, including human-operated collection through teleoperation, VR, or play~\cite{mees2022calvin,liu2023libero}, rule-based pipelines using privileged states, task logic, motion planning, or IK~\cite{james2020rlbench,gu2023maniskill2}, trajectory replay and augmentation from seed demonstrations~\cite{mandlekar2023mimicgen,jiang2024dexmimicen}, and autonomous or generative methods based on learned models and LLM/VLM planning~\cite{Wang2024robogen,chen2025robotwin}.

Recent work begins to study demonstration generation for dynamic manipulation. DynamicVLA~\cite{xie2026dynamicvla} uses manually designed state machines to collect dynamic data. DynaMimicGen~\cite{pomponi2026dynamimicgen} updates motion targets online with dynamic movement primitives (DMPs). DynamicManip~\cite{liao2026dynamicmanip} starts from a single static demonstration and synthesizes dynamic demonstrations by combining dynamic goal planning, anchor transformations for replaying local contact trajectories, and planned transitions. 
DOMINO~\cite{fang2026towards} explores large-scale dynamic manipulation data generation through synchronized replay, enabling rigid-body dynamics upon gripper–object contact. However, robot motions are replayed alongside predefined object trajectories rather than adjusted online to reactively track moving targets, limiting reactive robot–object interactions.
Overall, existing dynamic demonstration generation still relies mainly on state machines, target updates, or trajectory reconstruction, and lacks mechanisms that learn execution-level corrections near contact from rollout outcomes.

\subsection{Dynamic Object Manipulation}

Dynamic object manipulation remains a challenging problem in robot manipulation. Early studies focus on specific tasks, including dynamic grasping~\cite{Morrison-RSS-18}, throwing~\cite{zeng2019tossingbot}, table tennis~\cite{mulling2013learning,durr2026outplaying}, and catching~\cite{zhang2025catch,ren_catch_2026}, while DBC-TFP~\cite{zhang2025dynamic} and GEM~\cite{li2025sim} later explore data-driven dynamic manipulation. Recent work extends to more general dynamic scenarios. DynamicManip~\cite{liao2026dynamicmanip} detects task phases and adjusts execution parameters online. DynamicVLA~\cite{xie2026dynamicvla} uses continuous inference and latent-aware action streaming to reduce the mismatch between perception and execution. PUMA~\cite{fang2026towards} represents scene motion using historical optical flow and implicitly predicts future target states. PhysMani~\cite{yun2026physmani} explicitly models future 3D dynamics with a physics-constrained 3D Gaussian velocity field. DynamicWAM~\cite{lou2026dynamicwam} further combines historical optical flow with kinematic descriptions such as displacement, velocity, and acceleration, and adopts asynchronous execution to improve responsiveness. These methods mainly improve policies through dynamic representation, future prediction, and real-time execution. However, their data acquisition still depends heavily on task-specific experts or predefined motion patterns, and a general demonstration-generation mechanism that can autonomously adapt to diverse dynamic perturbations remains lacking.

\subsection{Residual Learning over Priors}

Reinforcement learning (RL) can improve robotic manipulation policies through online interaction, but direct RL often suffers from inefficient exploration, sparse rewards, and extensive reward engineering~\cite{bai2025towards}. Residual learning over priors therefore provides a promising alternative. Residual Policy Learning learns residual actions on top of priors such as planners and model predictive control (MPC), while CR-DAgger extracts incremental actions from online human corrections~\cite{silver2018residual,xu2026compliant}. ResiP~\cite{ankile2025imitation}, Policy Decorator~\cite{yuan2025policy}, and the recent DICE-RL~\cite{sun2026prior} further perform residual optimization over pretrained policies, focusing exploration on distribution shifts and execution-level errors that are not covered by the prior. At the algorithmic level, unlike PPO~\cite{schulman2017proximal}, which uses a critic to estimate values across states, methods such as GRPO~\cite{shao2024deepseekmath} and DAPO~\cite{yu2026dapo} compare multiple rollouts under the same condition. By holding difficult-to-model environmental factors fixed within each group, they concentrate the learning signal on execution differences that actually change task outcomes. Group-relative objectives of this form are increasingly adopted in robot learning~\cite{li2026simplevla,Xue_2026_CVPR}. Following this principle, DynaForge learns execution-level residual corrections over a planner prior and constructs an implicit curriculum on top of GRPO.
\section{Preliminaries}
\label{sec:preliminaries}

\begin{figure*}[tbp]
    \centering
    \includegraphics[width=\textwidth]{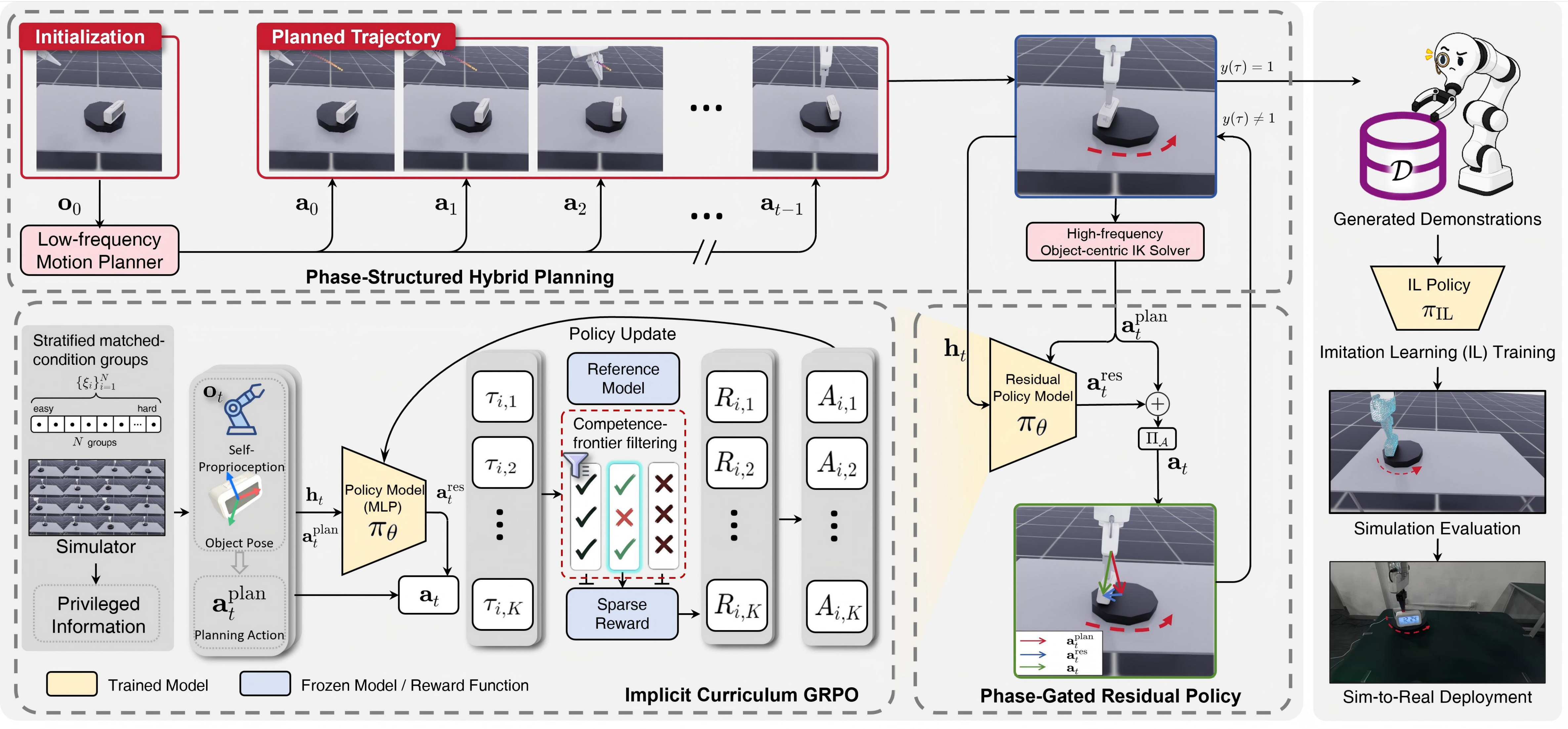}
    \caption{\textbf{DynaForge overview.} 
    A low-frequency motion planner provides nominal trajectories in geometrically stable phases, while high-frequency object-centric IK updates the planning action near dynamic contact boundaries and a residual policy corrects it before execution. For residual learning, rollouts under matched difficulty conditions form groups, and only mixed-success groups contribute to group-relative updates. Successful rollouts are retained as demonstrations for downstream imitation learning, simulation evaluation, and sim-to-real deployment.
    }
    \label{fig:dynaforge_overview}
\end{figure*}

\noindent\textbf{Conditional demonstration generation.}
Each task follows a phase sequence. The executing phase index $\ell_t$
advances when its success predicate is satisfied. A dynamic condition $\xi$
specifies the scene initialization and exogenous object dynamics. We model the
task as an episodic Markov Decision Process (MDP) $\mathcal{M}_{\xi}=(\mathcal{S},\mathcal{A},P_{\xi},r)$,
with simulator state space $\mathcal{S}$, action space $\mathcal{A}$,
condition-dependent transition $P_{\xi}$, and sparse task reward $r$.
With the phase program and planner fixed, the residual policy $\pi_{\theta}$ induces rollouts
\begin{equation}
    \tau\sim p(\tau\mid\xi,\pi_{\theta}).
    \label{eq:conditional_trajectory}
\end{equation}
In addition, the terminal success is defined as $y(\tau)=\mathbf{1}\!\left[\operatorname{Succ}(\tau)\right]$, where the success condition $\operatorname{Succ}(\cdot)$ determines the success of each rollout. 
The successful rollout are saved as the demonstration data as
\begin{equation}
    \mathcal{D}
    =\left\{(\mathbf{o}_{0:T},\mathbf{a}_{0:T})
      \;\middle|\; y(\tau)=1\right\},
    \label{eq:demonstration_set}
\end{equation}
where $\mathbf{o}_{0:T}$ and $\mathbf{a}_{0:T}$ are the observations and actions for the imitation learner.

\noindent\textbf{Object-centric geometry.}
We denote homogeneous transformations by $\mathbf{T}$. Let $O$ be the reference object and $E$ the end effector. For the active phase $\ell_t$, the desired end-effector pose in the world frame is
\begin{equation}
    {}^{W}\!\mathbf{T}^{\ell_t}_{E,t}
    ={}^{W}\!\mathbf{T}_{O,t}\,{}^{O}\!\mathbf{T}^{\ell_t}_{E},
    \label{eq:object_centric_target}
\end{equation}
where ${}^{O}\!\mathbf{T}^{\ell_t}_{E}$ is the corresponding target pose expressed in the object frame, and ${}^{W}\!\mathbf{T}_{O,t}$ is the object pose in the world frame and varies with the object movement.

\section{DynaForge}
\label{sec:method}

\subsection{Framework Overview}
As illustrated in~\figref{fig:dynaforge_overview}, DynaForge uses a phase-structured planner to provide nominal actions and learns residual corrections for dynamic interaction. Stable phases use low-frequency global planning, while moving or contact-sensitive phases use high-frequency object-centric inverse kinematics. A phase-gated residual policy is trained for correction in the dynamic phase. We further propose implicit curriculum group-relative learning to make the residual learning focusing optimization near the evolving competence frontier. During demonstration generation, the successful rollouts are retained for downstream imitation learning.

\subsection{Phase-Structured Hybrid Planning}
\label{sec:planning}
To balance global geometric feasibility with responsiveness to dynamic interactions, DynaForge adopts a phase-structured hybrid strategy for the planner. 
The phase program assigns each phase an object-centric target, a success predicate, a control mode, and a residual gate. 
This makes the task-specific structure used for generation. 
Stable free-space transitions use a global trajectory, whereas moving or contact-sensitive phases update a local target online.

\noindent\textbf{Low-frequency motion planning.}
At the entry to a stable free-space phase, DynaForge compute once the phase target through Eq.~\eqref{eq:object_centric_target}. The motion planner~\cite{sundaralingam2023curobo}
then computes a collision-aware trajectory, which is executed
until the phase predicate is satisfied. Global replanning therefore occurs at
phase transitions rather than at every control step.

\noindent\textbf{High-frequency object-centric IK.}
For a moving target or contact-sensitive phase, DynaForge compute the phase target through Eq.~\eqref{eq:object_centric_target} from the latest object pose and solves IK at each control step.

\subsection{Phase-Gated Residual Policy}
\label{sec:residual}

Online IK resolves the current geometric target but does not account for all timing and interaction errors near contact. We therefore learn a residual policy to correct the planner action during these contact-sensitive phases.
Let $\mathbf{a}^{\mathrm{plan}}_t$
be the planning action and $\mathbf{h}_t$ a short history of privileged robot and object state. During training, a residual action is sampled from the behavior policy:
\begin{equation}
    \mathbf{a}^{\mathrm{res}}_t
    \sim \pi_{\theta_{\mathrm{old}}}
      (\cdot\mid\mathbf{h}_t,\mathbf{a}^{\mathrm{plan}}_t),
    \label{eq:residual_action}
\end{equation}
where $\pi_{\theta_{\mathrm{old}}}
      (\cdot\mid\mathbf{h}_t,\mathbf{a}^{\mathrm{plan}}_t)$ is parameterized as the Gaussian distribution $\mathcal{N}\!\left(
        \boldsymbol{\mu}_{\theta_{\mathrm{old}}}
        (\mathbf{h}_t,\mathbf{a}^{\mathrm{plan}}_t),
        \sigma^2\mathbf{I}\right)$ for exploration.
The residual action is added to the planning action to generate the action:
\begin{equation}
\mathbf{a}_t
    =\Pi_{\mathcal{A}}\!\left[
        \mathbf{a}^{\mathrm{plan}}_t+m_{\ell_t}\mathbf{a}^{\mathrm{res}}_t
      \right],
\end{equation}
where $\Pi_{\mathcal{A}}$ enforces the action limits and $m_{\ell_t}\in\{0,1\}$ enables residual correction only in phases needing dynamic response.
At evaluation and
demonstration generation, exploration is removed, \emph{i.e.}, 
$\mathbf{a}^{\mathrm{res}}_t=\boldsymbol{\mu}_{\theta}
(\mathbf{h}_t,\mathbf{a}^{\mathrm{plan}}_t)$. 

\subsection{Implicit Curriculum Group-Relative Learning}
\label{sec:frontier_learning}

With independently sampled dynamic conditions, vanilla GRPO can conflate task difficulty with the quality of residual corrections. We therefore introduce an implicit curriculum for GRPO, using selected success--failure contrasts under matched conditions to focus learning on the policy's evolving competence frontier.

\noindent\textbf{Stratified matched-condition groups.}
Task difficulty is parameterized by quantities such as object speed or
perturbation magnitude. To cover the configured range within each update, we
partition it into $N$ strata of nominal difficulty and assign $K$ environments
to each stratum. For group $i\in\{1,\ldots,N\}$, we sample one condition $\xi_i$
from its stratum and share its initialization and perturbation parameters
across all $K$ environments. Residual actions are sampled independently,
yielding the sampled trajectories $\tau_{i,k}\sim p(\tau\mid\xi_i,\pi_{\theta_{\mathrm{old}}})$ for
$k\in\{1,\ldots,K\}$.

Residual learning uses a sparse episodic reward evaluated once per rollout.
Let $y_{i,k}=y(\tau_{i,k})$ denote terminal success and
$c_{i,k}\in\{0,\ldots,C\}$ the index of the furthest validated checkpoint among
$C$ ordered task-progress checkpoints. Without dense per-step shaping, the
return for environment $(i,k)$ is
\begin{equation}
    R_{i,k}=y_{i,k}+(1-y_{i,k})\frac{c_{i,k}}{C}.
    \label{eq:episodic_return}
\end{equation}
The group mean return and group-relative advantage are
\begin{equation}
    \bar R_i=\frac{1}{K}\sum_{k=1}^{K}R_{i,k},\qquad
    A_{i,k}=\frac{R_{i,k}-\bar R_i}
    {\operatorname{std}_{k}(R_{i,k})+\varepsilon},
    \label{eq:group_advantage}
\end{equation}
where $\operatorname{std}_{k}(\cdot)$ is the within-group standard deviation
and $\varepsilon$ is a small positive constant.

\begin{figure}[t]
    \centering
    \includegraphics[width=\columnwidth]{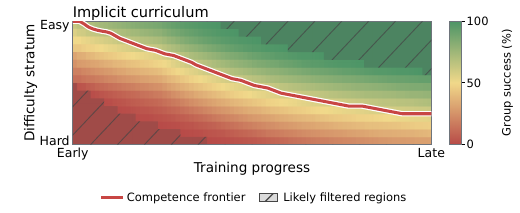}
    \caption{\textbf{Implicit curriculum over training.} Shading marks regions likely to yield homogeneous groups that are filtered. The red line illustrates the competence frontier advancing from easy to
    hard strata.}
    \label{fig:frontier_insight}
\end{figure}

\noindent\textbf{Competence-frontier filtering.}
All-failure groups provide no terminal-success contrast, even when partial
progress differs, while all-success groups expose no failure to correct.
Therefore, to focus the gradient signal on the competence frontier and adaptively selects the conditions that
contribute to optimization, we define the frontier
filter $\mathcal{F}$ to retain groups that contains both terminal successes and failures:
\begin{equation}
    \mathcal{F}=\left\{i:0<\sum_{k=1}^{K}y_{i,k}<K\right\}.
    \label{eq:frontier_set}
\end{equation}
\figref{fig:frontier_insight} illustrates the competence frontier moving
toward harder strata as group success improves during training. Shading marks
regions likely to yield all-success or all-failure groups. The frontier filter
$\mathcal{F}$ excludes these homogeneous groups from gradient computation and
skips the optimizer update when $\mathcal{F}=\varnothing$. As competence
improves, updates shift toward harder mixed-success conditions, forming an
implicit curriculum without a prescribed difficulty schedule.

\noindent\textbf{Group-relative policy update.}
We optimize the residual policy with the frontier-filtered GRPO
loss~\cite{shao2024deepseekmath}:
\begin{equation}
\begin{aligned}
    \rho_{i,k,t}(\theta)
    &=\frac{\pi_{\theta}(\mathbf{a}^{\mathrm{res}}_{i,k,t})}
    {\pi_{\theta_{\mathrm{old}}}(\mathbf{a}^{\mathrm{res}}_{i,k,t})},\\[-1pt]
    \widetilde{\rho}_{i,k,t}(\theta)
    &=\operatorname{clip}\!\left(
      \rho_{i,k,t}(\theta),1-\epsilon_{\mathrm{low}},
      1+\epsilon_{\mathrm{high}}\right),\\[-1pt]
    \mathcal{L}_{\mathrm{DF}}(\theta)
    &=-\mathbb{E}_{\mathcal{I}_{\mathcal{F}}}
    \!\left[\min\!\left(
      \rho_{i,k,t}(\theta)A_{i,k},
      \widetilde{\rho}_{i,k,t}(\theta)A_{i,k}\right)\right].
\end{aligned}
    \label{eq:frontier_grpo}
\end{equation}
Here $\rho_{i,k,t}$ is the likelihood ratio between the current and behavior
policies, with the common conditioning in Eq.~\eqref{eq:residual_action}
omitted for brevity. The operator $\operatorname{clip}(\cdot)$ truncates this
ratio to $[1-\epsilon_{\mathrm{low}},1+\epsilon_{\mathrm{high}}]$, yielding
$\widetilde{\rho}_{i,k,t}$. The set $\mathcal{I}_{\mathcal{F}}$ contains
residual-enabled transitions from selected groups, with the episode-level
advantage $A_{i,k}$ shared across their time steps. We adopt DAPO's asymmetric
clipping~\cite{yu2026dapo,li2026simplevla}, setting
$\epsilon_{\mathrm{low}}=0.20$ and $\epsilon_{\mathrm{high}}=0.28$ to allow a
wider upper clipping range for positive-advantage samples.

\section{Experiments}
\label{sec:experiments}

\begin{table*}[htbp]
    \caption{Matched dynamic demonstration generation.}
    \label{tab:dgr_results}
    \centering
    \footnotesize
    \newcommand{\DGRstd}[1]{\nobreak\hspace{0.08em}{\scriptsize\ensuremath{\pm}#1}}
    \setlength{\tabcolsep}{1.4pt}
    \renewcommand{\arraystretch}{1.25}
    \begin{tabular}{@{}>{\scriptsize}l*{9}{c}>{\columncolor{black!4}}c@{}}
        \toprule
        \rowcolor{black!4}
        \textbf{Generator} & Can & Bottle & Toy car & Lemon & Alarm Clock
        & Peach & Block & Pen & Peg & \textbf{Mean} \\
        \cmidrule(lr){2-10}\cmidrule(l){11-11}
        \textit{Dominant skill} & {\scriptsize grasp}& {\scriptsize grasp}& {\scriptsize grasp}& {\scriptsize catch}& {\scriptsize click}& {\scriptsize grasp + place}& {\scriptsize grasp + place}& {\scriptsize insert}& {\scriptsize grasp + insert}& -- \\
        \midrule
        DynamicVLA
        & 42.67\DGRstd{5.44} & \underline{28.00}\DGRstd{2.16} & 51.67\DGRstd{2.49} & N/A & N/A
        & \underline{19.33}\DGRstd{4.99} & 3.67\DGRstd{2.49} & N/A & N/A & 29.07\DGRstd{0.50} \\
        DMG
        & 20.67\DGRstd{1.25} & 12.67\DGRstd{4.19} & 16.67\DGRstd{1.70} & 14.33\DGRstd{2.05}
        & 7.33\DGRstd{3.86} & 9.00\DGRstd{3.27} & 12.33\DGRstd{2.05} & 27.67\DGRstd{1.70}
        & 0.00\DGRstd{0.00} & 13.41\DGRstd{1.48} \\
        \midrule
        DynaForge (w/o residual policy)
        & \underline{46.67}\DGRstd{2.05} & 21.33\DGRstd{1.25} & \underline{67.67}\DGRstd{2.62} & \underline{72.00}\DGRstd{2.45}
        & \underline{51.33}\DGRstd{1.70} & 18.67\DGRstd{4.50} & \underline{54.00}\DGRstd{3.74} & \underline{32.33}\DGRstd{2.49}
        & \underline{7.67}\DGRstd{0.94} & \underline{41.30}\DGRstd{1.00} \\
        \rowcolor{MethodGreen!9}
        \textbf{DynaForge (w/ residual policy)}
        & \textbf{74.67}\DGRstd{2.05} & \textbf{79.00}\DGRstd{2.16}
        & \textbf{98.67}\DGRstd{1.25} & \textbf{84.00}\DGRstd{2.16}
        & \textbf{100.00}\DGRstd{0.00} & \textbf{75.00}\DGRstd{2.94}
        & \textbf{96.67}\DGRstd{1.70} & \textbf{69.33}\DGRstd{1.25}
        & \textbf{28.00}\DGRstd{3.74} & \cellcolor{MethodGreen!16}\textbf{78.37}\DGRstd{1.41} \\
        \midrule
        \textit{$\Delta$ (pp)}
        & +28.00 & +57.67 & +31.00 & +12.00 & +48.67 & +56.33
        & +42.67 & +37.00 & +20.33 & +37.07 \\
        \bottomrule
    \end{tabular}
    \par\smallskip
    \begin{minipage}{\textwidth}
    \footnotesize\raggedright
    \textit{Notes.} Data-generation rate
    (DGR, \%; mean $\pm$ standard deviation over three seeds, each
    with 100 rollouts). The skill row records each task's dominant interaction.
    $\Delta$ is the difference between the reported DynaForge means in percentage
    points. N/A denotes a task outside a baseline's supported skill set. The
    DynamicVLA mean averages its five supported tasks; all other means average
    the nine tasks.
    Best means are \textbf{bold}; second-best means are \underline{underlined}.
    \end{minipage}
\end{table*}

\begin{figure}[!t]
    \centering
    \includegraphics[width=\columnwidth]{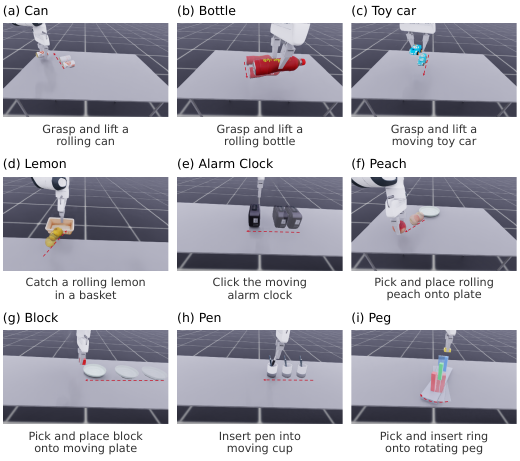}
    \caption{\textbf{Simulation dynamic manipulation tasks.} Panels (a)--(i) show each task and its objective, and red arrows indicate motion patterns.}
    \label{fig:benchmark_montage}
\end{figure}

Our experiments address four questions:

\noindent\textbf{Q1: Demonstration yield and collection cost.} How reliably does DynaForge generate successful demonstrations across dynamic manipulation skills, and how does its preparation cost compare with existing generators?

\noindent\textbf{Q2: Downstream data utility.} Do DynaForge demonstrations improve simulation policy success compared with DOMINO data under the same dataset size and training budget?

\noindent\textbf{Q3: Sim-to-real transfer.} How well do policies trained on DynaForge demonstrations transfer to physical dynamic grasping, clicking, and placement compared with policies trained on DOMINO data?

\noindent\textbf{Q4: Implicit curriculum and training efficiency.} How does the competence frontier evolve during learning, and how does the curriculum affect final success, optimizer steps, and training time relative to vanilla GRPO?

\subsection{Experimental Setup}
\label{sec:experimental_setup}

\noindent\textbf{Simulation and benchmark.}
We use Isaac Lab, built on Isaac Sim, with a Franka Research 3 arm and assets from RoboTwin and DOM~\cite{chen2025robotwin,fang2026towards}. The nine tasks cover dynamic grasping, catching, clicking, placing, and insertion, with rolling objects, translating receptacles, and rotating fixtures (\figref{fig:benchmark_montage}). Motion ranges from near-static conditions to launch speeds of approximately 1.5~m/s. Rolling objects evolve under simulated contact dynamics after initialization, while driven targets retain task-specific motion constraints. Training uses NVIDIA GeForce RTX 5090 GPUs.

\noindent\textbf{Robot platform.}
As shown in \figref{fig:sim2real_setup}, our robot platform consists of a 7-DoF Franka Research 3 arm and an external ZED RGB-D camera.

\noindent\textbf{Protocol and metrics.}
Data-generation rate (DGR) is the fraction of attempted rollouts that yield successful demonstrations, while downstream success rate (SR) measures task completion. Simulation comparisons use common task conditions and environment-defined success predicates, counting failures and timeouts as unsuccessful attempts. Unless stated otherwise, simulation results report mean and standard deviation over three evaluation seeds with 100 rollouts each. Physical evaluation uses ten trials per task and data source.

\subsection{Dynamic Demonstration Generation (Q1)}
\label{sec:exp_generation}

We compare DynaForge with DynamicVLA's state-machine collector~\cite{xie2026dynamicvla}, DynaMimicGen (DMG)~\cite{pomponi2026dynamimicgen}, and our planning prior without residual correction. 
Notably, DynamicVLA's collector relies on an authored state machine~\cite{xie2026dynamicvla}, making its task coverage dependent on task-specific rules and engineering choices. Extending the collector to additional skills therefore requires additional expert design. In our benchmark, we evaluate DynamicVLA on the five tasks for which the corresponding state-machine logic is supported, while the other four tasks are reported as N/A.
The residual comparison changes only whether correction is enabled, retaining the task program and planning prior.

As shown in \tabref{tab:dgr_results}, DynaForge improves the DGR of the planning prior on all nine tasks, raising the mean from 41.30$\pm$1.00\% to 78.37$\pm$1.41\%, with gains ranging from 12.00 percentage points on Lemon to 57.67 percentage points on Bottle. Across the baselines, DynamicVLA achieves a mean DGR of 29.07\% over its five supported tasks, while DMG achieves 13.41\% over all nine tasks. DynamicVLA relies on authored state-machine logic~\cite{xie2026dynamicvla}, while DMG updates DMP goals online but retains the source gripper commands~\cite{pomponi2026dynamimicgen}. In contrast, DynaForge retains task structure while learning execution-level corrections from rollout outcomes. Together, the gains over both the planning prior and existing baselines support the effectiveness of learned execution correction, although aggregate DGR does not isolate individual failure causes.

\begin{table}[t]
    \caption{Budgeted human effort (minutes) and estimated machine time (hours) across five jointly supported tasks.}
    \label{tab:generation_time}
    \centering
    \footnotesize
    \setlength{\tabcolsep}{5pt}
    \renewcommand{\arraystretch}{1.10}
    \begin{tabular}{@{}lcc@{}}
        \toprule
        Data preparation & Human (min) & Machine (h) \\
        \midrule
        DynamicVLA & $\approx 100$ & 26.9 \\
        DMG & $\approx 52.5$ & 25.4 \\
        Planner only & $\approx 75$ & 10.4 \\
        \midrule
        DynaForge: task setup & $\approx 50$ & -- \\
        DynaForge: residual training & -- & 2.5 \\
        DynaForge: collection & -- & 4.0 \\
        \midrule
        \textbf{DynaForge: total} & $\approx 50$ & \textbf{6.5} \\
        \bottomrule
    \end{tabular}
    \par\smallskip
    \begin{minipage}{\columnwidth}
    \footnotesize
    \textit{Notes.} Human budgets include 10 min/task for setup, plus 10 min/task of debugging for DynamicVLA, 5 min/task for planner-only, or 30 s/task for DMG seed collection~\cite{liao2026dynamicmanip}. These are budget assumptions, not measured times. Machine time excludes setup computation; dashes denote no separately reported time.
    \end{minipage}
\end{table}

\tabref{tab:generation_time} compares preparation costs on Can, Bottle, Toy car, Peach, and Block. Collection time is normalized to a 4~h DynaForge reference and scaled by the sum of inverse task-wise DGRs, assuming equal cost per attempt. Including 2.5~h of residual training gives a 6.5~h machine total, below the baseline estimates under this cost model.

\begin{figure}[!t]
    \centering
    \includegraphics[width=\columnwidth]{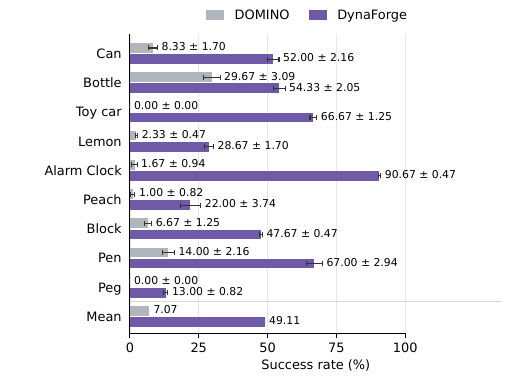}
    \caption{\textbf{DP3 success with 800 demonstrations per task.} Labels report mean $\pm$ standard deviation over three seeds of 100 rollouts, and error bars show the deviations. Mean averages nine task means.}
    \label{fig:simulation_results}
\end{figure}

\subsection{Downstream Policy Evaluation in Simulation (Q2)}
\label{sec:exp_simulation}

We train 3D Diffusion Policy (DP3)~\cite{Ze2024DP3} on 800 demonstrations per task generated by either DynaForge or DOMINO~\cite{fang2026towards}, using identical representations, preprocessing, 100,000 training updates, and evaluation procedures. Each policy receives point clouds and proprioception as input. By fixing the dataset size, we evaluate demonstration utility independently of generation yield. We use DOMINO as the downstream data-source baseline to compare demonstrations produced by the complete generation pipelines. 
We use DOMINO as the downstream data-source baseline, retaining its original synchronized replay and contact-handling mechanisms, to compare demonstration utility under matched downstream training budgets.

As shown in \figref{fig:simulation_results}, DP3 policies trained on DynaForge demonstrations outperform those trained on DOMINO data across all nine tasks, achieving a mean SR of 49.11\% compared with 7.07\%. The largest improvements are observed on Alarm Clock and Toy car, while Peg remains challenging for both methods. 
DOMINO generates demonstrations through synchronized replay of robot motions and predefined object trajectories~\cite{fang2026towards}, whereas DynaForge continuously adjusts robot actions to track moving objects and applies learned execution corrections. These results demonstrate the higher downstream utility of DynaForge demonstrations capturing reactive robot–object interactions.

\begin{figure}[!t]
    \centering
    \includegraphics[width=\columnwidth]{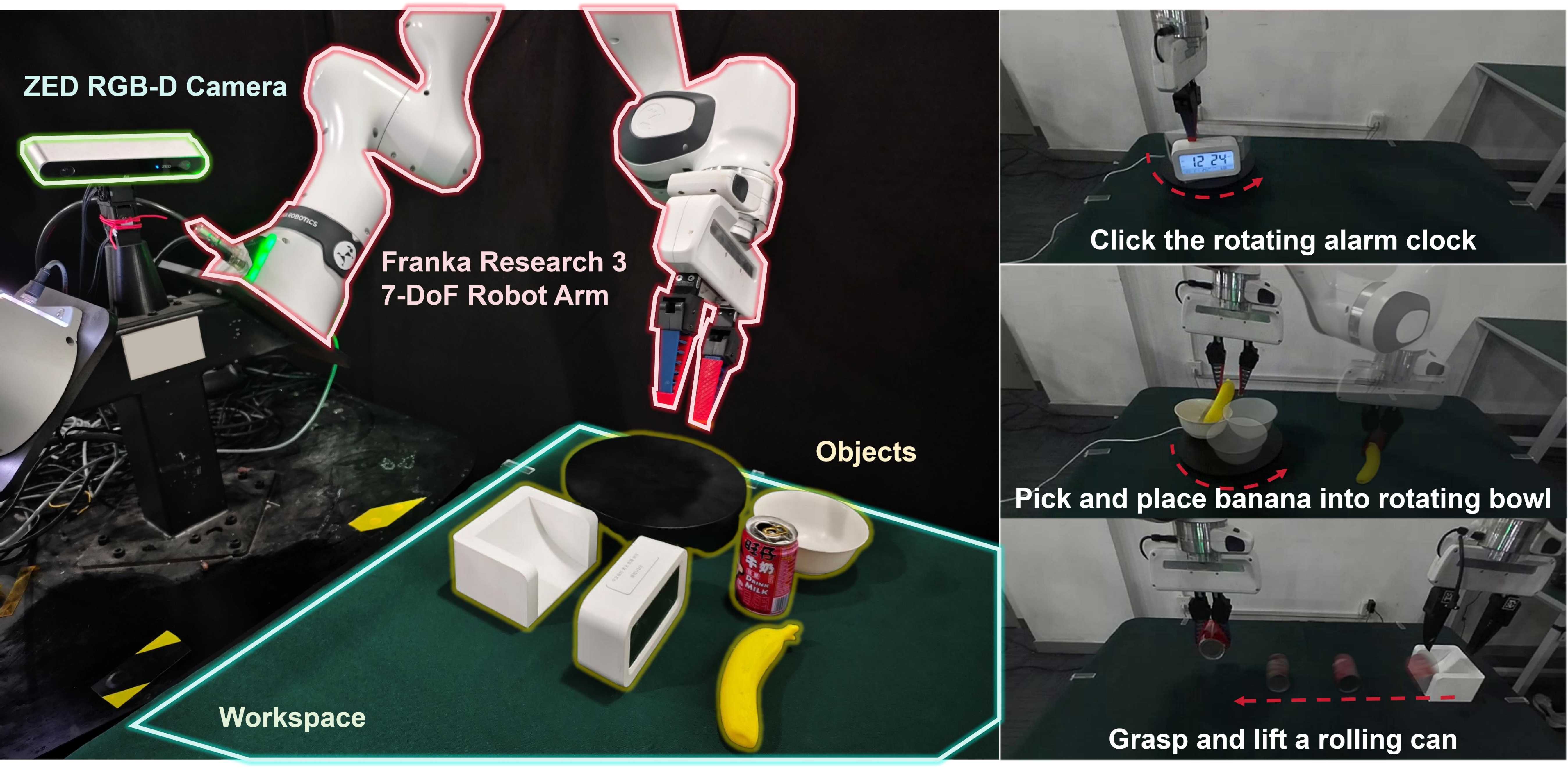}
    \caption{\textbf{Robot setup and tasks}. \textbf{Left} shows the Franka arm, ZED RGB-D camera, workspace, and task objects. \textbf{Right} shows the tasks including pressing a rotating alarm clock, placing a banana into a rotating bowl, and grasping a rolling can from top to bottom. Arrows indicate motion direction.}
    \label{fig:sim2real_setup}
\end{figure}

\begin{figure*}[!t]
    \centering
    \includegraphics[width=\textwidth]{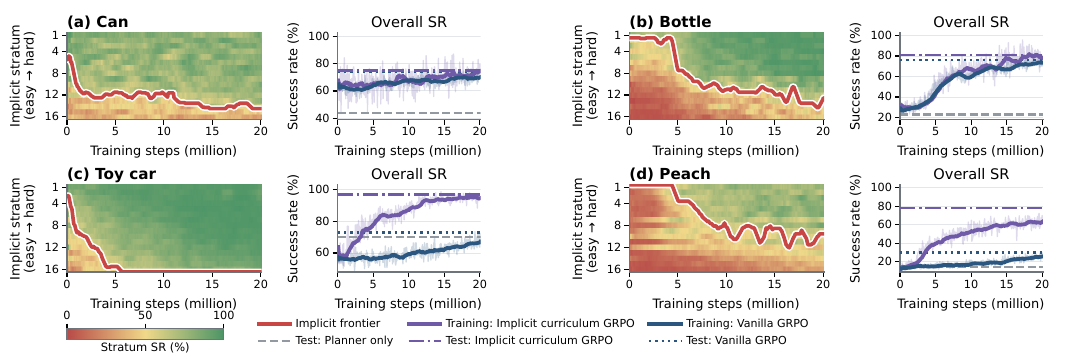}
    \caption{Evolution of the competence frontier and residual-training success. Each task pairs a stratum-success heatmap with training curves for implicit-curriculum and vanilla GRPO. Heatmaps and solid curves show trailing 20-update means; faint traces show per-update outcomes. Strata are ranked by full-run mean success for visualization only. The red frontier tracks the deepest contiguous rank whose 95\% Wilson lower bound reaches 50\%, with trailing nine-update smoothing. Horizontal lines show separate planner-only and final residual evaluations.}
    \label{fig:curriculum_ablation}
\end{figure*}

\subsection{Sim-to-Real Transfer (Q3)}
\label{sec:exp_sim2real}

As shown in \figref{fig:sim2real_setup}, we compare simulation-trained DP3 policies from the two data sources on three physical tasks: grasping and lifting a can rolling from right to left, pressing a rotating alarm clock, and placing a banana into a bowl on a rotating turntable. These tasks test dynamic grasping, clicking, and placement.

\begin{table}[t]
    \caption{Physical task success, reported as successful trials out of ten for each training-data source. Best results are bold.}
    \label{tab:sim2real_results}
    \centering
    \small
    \setlength{\tabcolsep}{4pt}
    \begin{tabular}{@{}lcc@{}}
        \toprule
        Task & DynaForge & DOMINO \\
        \midrule
        Grasp rolling can & \textbf{3/10} & 0/10 \\
        Click rotating alarm clock & \textbf{6/10} & 0/10 \\
        Place banana in rotating bowl & \textbf{5/10} & 1/10 \\
        \bottomrule
    \end{tabular}
    \vspace{-3mm}
\end{table}

As shown in \tabref{tab:sim2real_results}, DynaForge-trained policies achieve 30\%, 60\%, and 50\% success on dynamic grasping, clicking, and placement, compared with 0\%, 0\%, and 10\% for DOMINO. These results show that policies learn dynamic interaction skills from DynaForge demonstrations generated in simulation and transfer them to real world.

\subsection{Implicit-Curriculum Ablation (Q4)}
\label{sec:exp_curriculum}

We compare the DGR of DynaForge without a residual policy and residual generators trained with vanilla GRPO or the implicit curriculum on four tasks. Vanilla GRPO uses all complete rollout groups without difficulty stratification, whereas the implicit curriculum admits only matched-condition groups containing both successes and failures to optimization.

To visualize the frontier in \figref{fig:curriculum_ablation}, we smooth each stratum's observed success fraction $p_i(u)$ over the most recent updates:
\begin{equation}
    \hat p_i(u)=\frac{1}{w_u}\sum_{v=u-w_u+1}^{u}p_i(v),
    \qquad w_u=\min(20,u).
    \label{eq:frontier_window}
\end{equation}
Here $u$ indexes training updates. The difficulty strata $i\in\{1,\ldots,N\}$ are ordered by full-run mean success for visualization. Using pooled outcomes within the same window, we define the frontier as the deepest contiguous rank for which every stratum up to that rank has a 95\% Wilson lower bound of at least 50\%. The displayed frontier is further smoothed over the most recent nine updates.

The heatmaps show success extending toward strata with lower overall success as training progresses, accompanied by an advancing competence frontier. This pattern reflects an implicit curriculum that adapts as the policy improves. On Toy car and Peach, the training curves also show faster and larger gains than vanilla GRPO, supporting the value of learning from selected matched conditions.

\begin{table}[!t]
    \caption{Implicit-curriculum ablation (DGR, \%).}
    \label{tab:curriculum_results}
    \centering
    \footnotesize
    \setlength{\tabcolsep}{3pt}
    \renewcommand{\arraystretch}{1.15}
    \begin{tabular*}{\columnwidth}{@{}l@{\extracolsep{\fill}}rrrrr@{}}
        \toprule
        Generator & Can & Bottle & Toy car & Peach & Mean \\
        \midrule
        Planner only        & 44 & 23 & 70 & 14 & 37.75 \\
        Vanilla GRPO        & 74 & 76 & 73 & 30 & 63.25 \\
        Implicit curriculum & \textbf{75} & \textbf{81} & \textbf{97} & \textbf{78} & \textbf{82.75} \\
        \bottomrule
    \end{tabular*}
    \par\smallskip
    \begin{minipage}{\columnwidth}
    \footnotesize
    \textit{Notes.} Each cell uses 100 rollouts, evaluated separately from \tabref{tab:dgr_results}. Mean averages the four tasks.
    \end{minipage}
    \vspace{-3mm}
\end{table}

As shown in \tabref{tab:curriculum_results}, the implicit curriculum improves final DGR over vanilla GRPO by 1 and 5 percentage points on Can and Bottle, and by 24 and 48 percentage points on Toy car and Peach. The gains are positive across all four tasks, although their magnitude varies.

In the paired efficiency comparison, the implicit curriculum uses $0.73\times$ the optimizer steps and $0.62\times$ the training time of vanilla GRPO under the same nominal environment-step budget, while achieving higher final DGR. The reduction in optimization work and training time indicates improved optimization efficiency.

\section{Conclusion}
\label{sec:conclusion}

DynaForge turns phase-structured motion planning into a prior for dynamic demonstration generation. A residual policy corrects execution under privileged simulation, and an implicit curriculum focuses learning on mixed-success conditions. Experiments show higher demonstration yield across nine simulation tasks and improved downstream DP3 performance, with initial transfer to three physical skills.

Generation still depends on authored task structure and simulator fidelity. Low success on precision insertion and modest physical performance limit the present scope. Future work targets less manual task specification, better coverage of contact variations, and broader evaluation under changes in physical dynamics and visual observations.

\bibliographystyle{IEEEtran}
\bibliography{references}

\end{document}